# TEACHING CONTINUITY IN ROBOTICS LABS IN THE TIME OF COVID AND BEYOND

**R.P. Salas**

*Brandeis University (UNITED STATES)*

## Abstract

This paper argues that training of future Roboticists and Robotics Engineers in Computer Science departments, requires the extensive direct work with real robots, and that this educational mission will be negatively impacted when access to robotics learning laboratories is curtailed. This is exactly the problem that Robotics Labs encountered in early 2020, at the start of the Covid pandemic. The paper then turns to the description of a remote/virtual robotics teaching laboratory and examines in detail what that would mean, what the benefits would be, and how it may be used. Part of this vision was implemented at our institution during 2020 and has been in constant use since then. The specific architecture and implementation, as far as it has been built, is described. The exciting insight in the conclusion is that the work that was encouraged and triggered by a pandemic seems to have very positive longer-term benefits of increasing access to robotics education, increasing the ability of any one institution to scale their robotics education greatly, and potentially do this while reducing costs.

## 1 INTRODUCTION

As we all remember, early in 2020 we were required, practically from one day to the next, to abandon our campuses. We had to re-imagine and translate our courses and programs from an in-person to a remote or online model. Certain facilities such as the robotics teaching laboratory were shuttered, and ongoing projects had to go on hibernation as we conceived of a new approach. Through a process of experimentation, prototyping and a study of previous work, we were able to get up and running on a cluster of inexpensive gaming computers, organized and managed with Kubernetes [2] [1]. This approach worked well for our class of 13 students. Based on this work, and looking forward to when we returned to campus, we conceived of what we are calling a remote/virtual learning laboratory which is contribution of this paper.

### 1.1 Baseline

To set the stage, we briefly review the technical challenges of teaching robotics at the undergraduate level. Based on our experience, our requirements for teaching Robotics were simple and fairly mainstream. This was the baseline before the pandemic and even then, it was a challenge. We require:

- Robot Operating System (ROS) running on Ubuntu. Like many robotics labs, we have standardized on this operating environment. This is a fairly steep requirement, and it leads in a sense to the following ones.
- Capacity to support all the students in a class. Our ability to meet our scaling goals for the curriculum is dependent on many things, including the capacity of our technical infrastructure.
- Minimal dependency on the operating system, age, speed, memory and free diskspace of students' computers. Students come to university with all kinds of computers, and it is imperative that the lab software work on underpowered and under resourced computers.

Over several years we have refined our approach. We tried dual booting, external solid-state (SSD) boot drives, the Windows Linux subsystem, and were able to overcome the obstacles to an extent. But the truth was that setting up the environment was an obstacle and a discouragement for all involved.

### 1.2 Virtual Robotics Lab

Once it became clear that we would have to teach and work with students remotely we were forced to find a new solution to meet all the requirements. As some time had passed, new approaches became possible. The current clustered environment meets all the baseline requirements as mentioned above.

The cluster provides a scalable way to offer private Linux environments to all students, with the complete ROS software stack, plus all the course scaffolding. This environment is accessible by web browser putting a minimum load on the students' computers. Robotics requires a variety of graphical tools, all of which can be accessed by virtual desktop. This is a fairly standard software stack and addresses each of the requirements above. It is highly effective, as far as it goes.

## 2  HANDS-ON

The pedagogical value of hands-on experiences is well documented. There is a famous quote from Aristotle that one comes across regularly: "For the things we have to learn before we can do them, we learn by doing them." It is interesting to see this quote in context:

> *Again, of all the things that come to us by nature we first acquire the potentiality and later exhibit the activity (this is plain in the case of the senses; for it was not by often seeing or often hearing that we got these senses, but on the contrary we had them before we used them, and did not come to have them by using them); but the virtues we get by first exercising them, as also happens in the case of the arts as well. For the things we have to learn before we can do them, we learn by doing them, e.g., men become builders by building and lyre players by playing the lyre; so too we become just by doing just acts, temperate by doing temperate acts, brave by doing brave acts. [2]*

Jumping forward millennia, the question of the role of hands-on work, and in particular of the Laboratory as a valid and valuable component of scientific pedagogy has been much studied and debated. In "Best Practices in Robotics Education" [4], Thomas strongly argues for the importance of hands-on work in Robotics, saying "Robotics is at its heart, a multi-disciplinary, hands-on field. It invites people to get their hands dirty, persevere through trial and error, and experience the thrill of success". In the same paper, C. Berry, an experienced Robotics educator argues for the importance for students to "get their hands on something, even if it's just to see a motor spin".

Interestingly, hands-on learning in laboratories has not universally been seen as positive. For example, from Hofstein [5] we learn that often students spend so much time on technical minutia required to 'get the experiment to work' that the chances for learning were curtailed. Hofstein cites Gunstone: "Gunstone wrote that students generally did not have time or opportunity to interact and reflect on central ideas in the laboratory since they are usually involved in technical activities with few opportunities to express their interpretation and beliefs about the meaning of their inquiry." We can personally attest to the validity of these observations as well.

The reality is that a simulated robot can never take the place of a real robot. As Nuenuert et al point out [5], algorithms which are proven out on the most advanced simulators will not generally work in the same way on an actual robot. Depending on the way the algorithm is designed, it might not even work the same way on two identical actual robots. Current work [7] explores introducing randomness and imperfection into the simulators so as to challenge the robustness of their algorithms.

In Real Robots Don't Drive Straight [8], Martin makes powerful arguments for developing the intuition of students about the sharp and important difference between the idealized imaginary world of mathematics and simulation, and the real world of friction, traction, weight and power that they will encounter when working with real robots. Rawat et al [9] state that experience with real robotics systems is one of the two most important things provided to undergraduate Robotics students (the other being a strong knowledge of the fundamentals.

Finally, our own first-hand experience has taught us the importance of developing confidence in dealing with mechanical, electrical and electronic components. The old saying "measure twice cut once" applied to robots becomes "double check the spec sheets before you attach the connector to the wrong pin". How hard should I push the plug in to make sure it makes a solid connection? How tightly should I turn the screw? All this knowledge, which is really muscle memory and ability, can only be obtained with hands-on learning. It is also essential to understand that when dealing with the physical world, that 1 + 1 might equal 2.00001. Data is noisy or wrong, everything is subjected to probabilistic errors, to concurrency and race conditions, even to variances in CPU or memory speed which leads to variation in performance. Even assuming the robot is assembled correctly, is working correctly, and the algorithms all check out, the student has to deal with the real physical world. The two motors with the same part number are not really identical. The two identical wheels are not exactly the same. The carpet has a bump, and the sun is glaring through a window and blinding the Lidar.

Based on this, we argue that (perhaps unlike most other domains of Computer Science), hands-on experience is essential to learning in robotics. Simulation does have an important role to play and can be conducted on individual workstations, but it is incomplete.

## 3 REMOTE ROBOTICS LABORATORY

There are two equally important critiques of teaching robotics solely with simulated robots and environments: a) learners are not touching and feeling the actual robots and b) the robots are simulated – not real! This leads to the notion of a remote laboratory.

A remote laboratory is a conventional laboratory configured to allow use from a distance. The robot could physically be located on a college campus, at a maker lab or similar space. The learners in turn could be at a different college or enterprise. In this scenario, there would necessarily be some actual robots, different types and models in a physical space. The space could be staffed or without staff. Staff would be beneficial to configure the space, replace batteries, and other activities that require a personal touch. But analogous to a virtual lab, the remote would be remotely controlled. Think of it as a laboratory with an API.

### 3.1 Previous Work in Remote Laboratories

There are many examples of remote robotic labs. Our review identifies some categories.

First, there are examples of remote-controlled robots meant to simply demonstrate the possibilities. For example, ReRobot Sandbox [10] is quite impressive. Various kinds of physical robots are shown on video and from a web page a user can play with a python program to make the robots move and act. This is certainly an existence proof, but it lacks support for writing and testing complex, concurrent, robotics algorithms and it also is not oriented around education.

Second, there are examples of true remote laboratories and factories in industry: Godfrey [11] describes a fully integrated, globally accessible, automated chemical synthesis laboratory. Amazon is famous for its fleet of remotely controlled robots, see for example [13]. These are impressive accomplishments which are advancing rapidly but of course are not oriented around education.

Third, there are examples of research labs that have created remotely controlled robot experiments. One of many examples is the Robotarium [15], a remotely accessible, multi-robot research facility. These are the closest but as they are single-purpose labs, very advanced, but oriented around research, and so very narrowly focused.

Fourth, there are very recent commercial offerings which are the closest to what this paper will describe. In particular we want to look at products from The Construct and Amazon. Amazon's product, RoboMaker [13] is a full featured cloud-based development environment for Robots. It is quite complex and quite expensive. The product from The Construct [11] is fairly similar to the present work. It is oriented around teaching, works fully in the cloud, and supports general robotics work.

In [16] Wästberg et al describe at length challenges with creating virtual laboratories in Physics education. They cite challenges in interaction design, visualization, engineering and pedagogy. They also discuss the role and limitations of simulation.

### 3.2 Remote/Virtual Robotics Laboratory

On the one hand, hands-on work is essential to learning robotics, and on the other hand, sometimes due to time, geography and sheer scale, having a virtual learning environment is the only option. This dilemma led us to remote/virtual robotics laboratories. To classify different approaches that may be taken, we offer the following definition of tiers of service:

1. **Tier 1:** Simulated Environments: These environments look like video games to our students and are immediately familiar and attractive. Picking up a robot in a simulated environment and dropping it on its side, with the wheels still spinning always produces laughter and draws the students closer. It's the best place to start.

2. **Tier 2:** Cheap Robots: Very simple robots are a suitable next step. They are inexpensive enough (like a pricey textbook) that we can consider supplying each student their own. However, their simplicity immediately is a ceiling that students will want to break through.

3. **Tier 3:** Remote Physical Lab: This would provide the greatest flexibility but is also the most complex. The fact that we could achieve remote access would mean that the lab could be used 24x7 and the lab could be located in less desirable and inexpensive space.

## 3.3 Key Requirements

Based on our research and experience we can now describe an effective remote/virtual robotics Laboratory. We believe the following are the key requirements:

1. **Accessibility:** The Lab should be accessible and usable from any kind of computer, running any operating system. Requiring a high-end computer would limit accessibility.
2. **Remote Use:** Provisioning and use should be possible without requiring physical presence. Students can get set up and productive without ever setting foot on campus.
3. **Programming Environment:** The lab should provide an easy programming environment where students can build on their knowledge and use well established, open and not proprietary tools, even as they move through the tiers.
4. **Scale:** The Lab should allow classes of as many as 100 students to use its resources (in Tier 1 and 2) simultaneously, and Tier 3 asynchronously.  In addition, scale requires that adding new students is an automated process, not requiring expert access to each student's computer.
5. **Simulation:** The Lab should provide access to simulated robots and worlds, to address Tier 1 of Robotics learning environments described earlier.
6. **Coordinated use of Lab robots:** The Lab should provide for remote "time-shared" access to physical robots in the Lab.
7. **Cost:** Additional cost to the student should not exceed the cost of an expensive textbook.

## 3.4 Challenges

It is apparent that our virtual robotics lab described above does meet some of the above requirements. Meeting all of them however presents numerous challenges, many firmly in the camp of "easy in theory, not so easy in practice."

Accessibility in the way described above immediately suggests delivering the Lab experience in a standard web browser. This has been seen for some years now in the form of increasingly sophisticated "web IDE" tools. In robotics there is another major level of challenge. The well-established simulation tools such as Gazebo are extremely resource hungry and graphics intensive. The recommended approach is to create a web-based environment which provides an IDE, a command line and a windowed capability for simulation, and hosting this on a GPU equipped node.

The scale requirement has two separate elements: 1) The ability to smoothly grow the number of students using the Lab without a hard maximum and 2) not requiring an expert to set up each student's computer. Both those challenges can be solved by starting with a "Robot as a Service" [13] mindset and using containerization and cloud computing instances.

Once we provide web-based access to a containerized node we are well on the way to a useful programming environment. The present reality is that programming, debugging and testing a robot is a complicated and messy process. For anything sophisticated we are not yet in a plug-and-play world. The software engineer will often have to step outside of the IDE to deal with proprietary tools and file formats, custom network configurations, IP addresses and similar complications.

The idealized vision for providing actual, inexpensive robots to students to use at their own location (Tier 2) requires a few key hurdles to be addressed: cost, assembly and test, provisioning and networking. At the present time (May 2021) a robot equipped for introductory and intermediate robotics students in a Computer Science Department - having a differential drive robot with Wi-Fi, LIDAR and a camera, running Linux and ROS, will cost around USD $270. This is still too expensive. In our experience, many undergraduate students are not confident in their ability to assemble and program an expensive robot without in person guidance.

In this final section we will actually outline an attempt at addressing all these needs. This is a work in progress, where some parts are solved, and many are not. However, the present system is currently deployed and in use at our institution.

# 4 CURRENT STATUS

Currently our virtual lab environment is in active use, reliable and repeatable. There were very few issues or complaints from students. But it is not yet easy enough to operate and manage. On the other hand, the remote-controlled physical lab is still on the drawing boards.

## 4.1 Virtual lab environment using containerization

Containerization is a well understood technology for deployment of a standardized configuration onto computers of different architectures. It is not a magic bullet, however. To accomplish the stated goals, it is necessary to create a configuration that includes both the graphical and non-graphical components and provides access from a remote web browser.

Once the containerization is solved, it is necessary to deploy the container on the available cloud hosts, and new nodes have to be provisioned (and paid for) and de-provisioned as students come and go. To meet the scalability goal, this process must be automated to allow students to be added and deleted, and their credentials tracked and communicated. Connectivity is achieved by defining a virtual private network (VPN) that spans all the cloud-based environments, the student's own desktop computers, their own robots, as well as Robots in the central lab.

## 4.2 Physical Robots

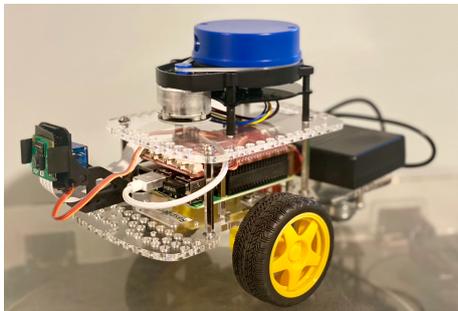

We offer inexpensive robots to students at cost, assembled and tested. We elected not to make them required because of the cost. In other words, a student can succeed in the course without having their own robot.

Our experience is that a new set of problems arises as students acquire their own robots. Connecting them to their Wi-Fi, having physical space at home to work with the robots and other practical problems troubleshooting, all interfere with their work. For the teachers, there are challenges installing, monitoring and updating the software on the robots, and when the software has to be updated it is not unusual to require a multi gigabyte file to be sent out to each robot, perhaps burned onto a new memory card, and then a reboot. Many students could handle this but certainly plenty would have trouble.

## 4.3 What Comes Next

The solutions outlined above are all part of the current implementation. This section reviews the missing pieces that will require to realize the overall vision of Remote access to a central lab with actual robots.

Recall the third Tier described early in this paper. It requires a central Lab, an actual physical lab space. Unlike typical labs however, this one does not need to be accessed by local students. In fact, there are various interesting and innovative ways to approach this. A centralized lab could be located off campus in less desirable real estate. Because students were visiting the lab in person, staffing could be sharply reduced. Heating and cooling expenses would be reduced. It could be shared among multiple institutions, or high schools. It could even be in another region or country.

The central lab as envisioned will require a combination of hardware, software and minimal staff. Part of the floorspace of the lab is set up as a field on which the robots will travel. It could have any shape, mazes, fake scenarios. In fact, users of the lab could request different fields to be set up. Actual robots will be placed on the field. They will be controlled indirectly by the remote users, but directly by a control server that is inside the lab. In addition, there would be multiple webcams to allow the robots to be seen and monitored by the remote users (students.) To help understanding, here's an example, currently imaginary, scenario.

A student uses the Lab's web site to allocate a time slot and one or robots that will be available at that time. When the time arrives, the system automatically re-provisions the robots to make sure their software is "reset to factory settings." The web site displays live video of the lab from more than one angle. At the same a docker image is brought up, configured with the IP addresses of the robots. The student can use the web-based IDE to write programs and test them and see the effects on the live robots.

## 5 CONCLUSION

While this work was initiated before the pandemic it received a major boost in attention and drive as education was moved remote and students could no longer come in person to the robotics lab at our institution. While our vision of a remote/virtual robotics learning laboratory has not been fully realized yet, what we have so far has been successfully used for almost two semesters.

With this experience we have come to realize that this vision, if fully realized, will profoundly increase the reach and access to this very exciting aspect of computer science and computer science education. Where previously we were limited by the size of our physical lab and geographical location of our students, the remote/virtual robotics learning laboratory promises to offer access to many more students in more locations at less cost.